\pdfoutput=1
\documentclass[11pt]{article}

\usepackage[]{ACL2023}

\usepackage{times}
\usepackage{latexsym}

\usepackage[T1]{fontenc}
\usepackage[utf8]{inputenc}

\usepackage{microtype}

\usepackage{inconsolata}

\usepackage{tabularx}
\usepackage{todonotes}
\usepackage{multirow}
\usepackage{graphicx}
\usepackage{enumitem}
\usepackage{anyfontsize}
\usepackage{url}
\usepackage{subcaption}
\usepackage{xr}
\usepackage{tabu}
\usepackage{bbold}
\usepackage{booktabs}
\usepackage{xcolor,pifont}
\usepackage{MnSymbol,wasysym}
\usepackage{fontawesome}
\usepackage{xspace}
\usepackage{caption}
\usepackage{array}
\usepackage{tcolorbox}
\usepackage{xcolor,colortbl}
\usepackage{makecell}
\usepackage{soul}
\usepackage{stackengine}
\newcolumntype{H}{>{\setbox0=\hbox\bgroup}c<{\egroup}@{}}
\newcommand{\vtwo}[1]{{\color{black}{#1}}}

\newcommand{\vfour}[1]{{\color{black}{#1}}}
\newcommand{\vfive}[1]{{\color{black}{#1}}}
\newcommand{\vfinal}[1]{{\color{black}{#1}}}

\newcommand{\Skip}[1]{}

\newcommand{\dataname}{\textsc{Maccrobat-EE}\xspace}
\newcommand{\sourcedata}{\textsc{Maccrobat}\xspace}
\newcommand{\modelname}{\textsc{Dice}\xspace}

\newcommand{\ie}{\textit{i}.\textit{e}.\ }
\newcommand{\eg}{\textit{e}.\textit{g}.\ }

\newcommand{\secref}[1]{\S\ref{#1}}
\newcommand{\appref}[1]{Appendix~\ref{#1}}
\newcommand{\figref}[1]{Figure~\ref{#1}}
\newcommand{\tbref}[1]{Table~\ref{#1}}

\newcommand{\dotieconcat}[2]{% auxiliary macro, don't use it directly
  \text{\raisebox{.8ex}{$\smallfrown$}}%
}
\newcommand{\mypar}[1]{\paragraph{#1}}
\definecolor{lightblue}{RGB}{212, 235, 255}
\definecolor{lightorange}{RGB}{255, 204, 168}
\definecolor{lightyellow}{RGB}{255, 255, 168}
\sethlcolor{lightblue}
\newcommand\hlc[2]{\sethlcolor{#1}\hl{#2}}
\newcommand{\vanillaDiceED}
{& 65.03 & 74.08 & 69.26 & 60.51 & 70.28 & 65.03}
\newcommand{\vanillaDiceEAEGold}
{& 49.10 & 53.60 & 51.25 & 45.95 & 50.76 & 48.24}
\newcommand{\vanillaDiceEAE}
{& 70.76 & 76.48 & 73.51 & 66.47 & 72.71 & 69.45}
\title{DICE: Data-Efficient Clinical Event Extraction with Generative Models}

\author{
Mingyu Derek Ma\Thanks{~Equal contribution.}~~~
Alexander K. Taylor\footnotemark[1]~~~
Wei Wang~~~
Nanyun Peng
  \\
  Computer Science Department
  \\
  University of California, Los Angeles
  \\
  \texttt{\{ma, ataylor2, weiwang, violetpeng\}@cs.ucla.edu}
}

\date{}

\begin{document}
\maketitle

\begin{abstract}
    Event extraction for the clinical domain is an under-explored research area. The lack of training data along with the high volume of domain-specific terminologies with vague entity boundaries makes the task especially challenging. In this paper, we introduce \modelname, a robust and data-efficient generative model for clinical event extraction. 
\modelname frames event extraction as a conditional generation problem and introduces a contrastive learning objective to accurately decide the boundaries of biomedical mentions. \modelname also trains an auxiliary mention identification task jointly with event extraction tasks to better identify entity mention boundaries, and further introduces special markers to incorporate identified entity mentions as trigger and argument candidates for their respective tasks. To benchmark clinical event extraction, we compose \dataname, the first clinical event extraction dataset with argument annotation, based on an existing clinical information extraction dataset, \sourcedata~\cite{maccrobat}. Our experiments demonstrate state-of-the-art performances of \modelname for clinical and news domain event extraction, especially under low data settings.
\end{abstract}

\section{Introduction}
\label{sec:intro}

Event extraction (EE) is an information extraction task that aims to identify event triggers and arguments from unstructured texts \cite{ahn-2006-stages}. The EE task consists of two subtasks: 1) event detection, in which the model extracts trigger text and predicts the event type; and 2) event argument extraction, in which the model extracts argument text and predicts the role of each argument given an event trigger and associated event type. 

\begin{figure}
    \centering
    \includegraphics[width=\columnwidth]{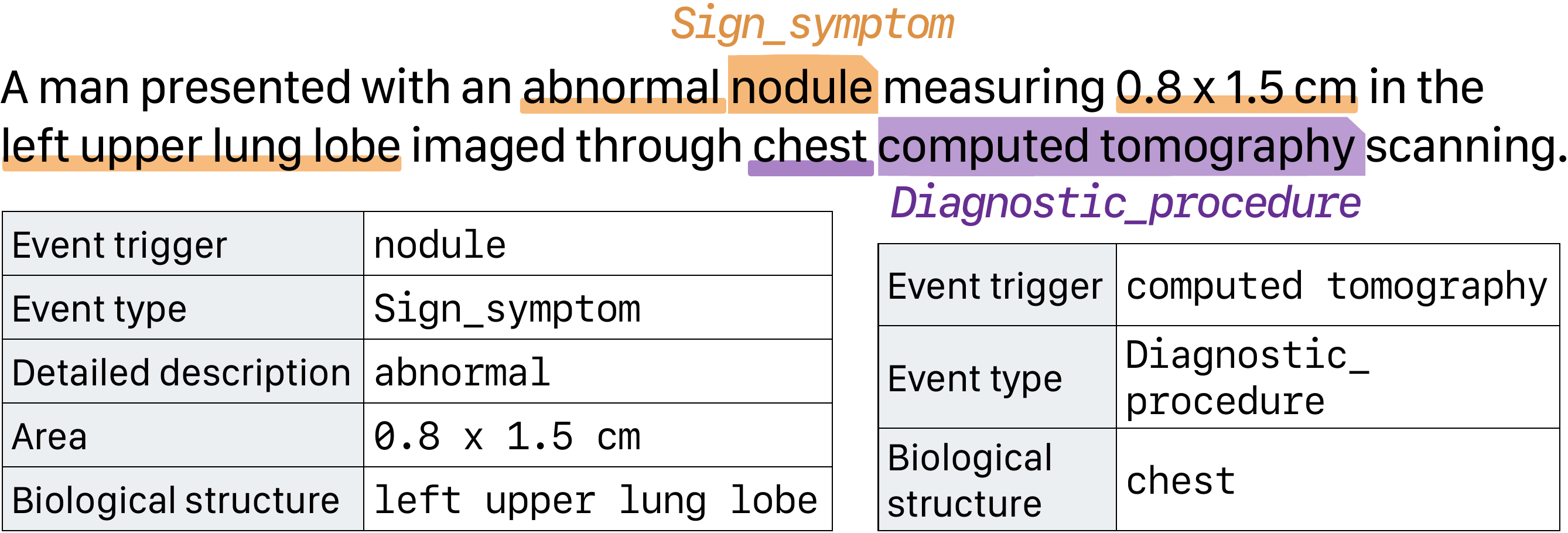}
    \caption{Illustration of a \textsc{Sign\_symptom} event triggered by ``nodule'' \vfive{with multiple arguments including an }\textsc{Area} argument ``0.8x1.5cm'',
    % \textsc{Detailed\_description} argument ``abnormal'' and \textsc{Biological\_structure} argument ``left upper lung lobe'',
    and a \textsc{Diagnostic\_procedure} event whose predicate is ``computed tomography'' described by argument ``chest'' of role \textsc{Biological\_structure}.\looseness=-1 
    }
    \vspace{-0.8em}
    \label{fig:teaser}
\end{figure}

Clinical EE aims to extract clinical events, which are occurrences at specific points in time during a clinical process, such as diagnostic procedures, symptoms, etc. The arguments for such events are entities that modify or describe properties of these events \cite{maccrobat}. \figref{fig:teaser} shows an example sentence with two clinical events.
The overwhelming volume and details of clinical information necessitate clinical EE, which benefits many downstream tasks such as adverse medical event detection~\cite{rochefort2015accuracy}, drug discovery~\cite{wang2009active}, clinical workflow optimization~\cite{hsu2016data},
% and the construction of patient histories to inform 
and automated clinical decision support~\cite{yadav2013automated}.

However, there are several non-trivial challenges of clinical EE compared to general domain EE. 
% Challenge: domain knowleddge in NL
% First, domain knowledge such as clinical term definitions and semantic meanings are mostly preserved in natural language; however, traditional information extraction models cannot easily leverage such information.
% Challenge: mention vague and long
First, most triggers and arguments of clinical events consist of domain-specific terms that are more than 50\% longer than the general domain on average, as shown in \tbref{table:data_stats}, and have vague boundaries because most clinical mentions\footnote{Clinical mentions are defined as meaningful text spans of occurrences or their properties \cite{maccrobat}.} contain several descriptors.
% Locating and bounding clinical mentions requires special model design and domain understanding. 
% MACCROBAT paper: events occur during specific points in time (i.e.,they may be arranged into a timeline) while entities are other meaningful text spans, often those modifying or describing properties of events.
For instance, given the text span ``massive heart attack'', ``heart attack'' should be identified as the trigger  (instead of ``massive heart attack'' or ``attack'') because it refers to a specific condition, and ``massive'' is an argument of the role type \textsc{Severity}. 
However, when we consider ``right common carotid artery'', the entire text span describes a biological structure, and thus it functions as an argument of the role type \textsc{Biological\_Structure} despite ``right'' and ``common'' being descriptors for ``carotid artery''. 
% Challenge: long tail arguments
The second challenge is the diversity and density of clinical arguments: 
there are on average 10 unique argument roles for each clinical event type compared to 3.7 in the general domain.
% Challenge: no data
Finally, it is challenging to obtain high-quality annotated data for clinical events due to both patient privacy concerns and the cost of expert annotations. Due to these challenges, there have been no clinical EE datasets with argument annotations to the best of our knowledge.

% Ours
In this paper, we present \modelname, a \textbf{D}ata-eff\textbf{I}cient generative model for \textbf{C}linical \textbf{E}vent extraction.\footnote{\vfinal{Please refer to \url{https://derek.ma/DICE} for code and data.}} We build upon existing prompt-based generative event extraction models to formulate EE as a sequence-to-sequence text generation task~\cite{hsu-etal-2022-degree,ma-etal-2023-star}. 
% \modelname incorporates several techniques to address the challenges introduced by clinical terminology, which, as shown in this work, generalize to other domains with lengthy technical terms and long-tail argument roles.
%This enables us to leverage event type definitions and argument role descriptions as prompts to incorporate domain knowledge and boost low-resource performance. 
To handle the special challenges of clinical EE, \modelname 1) introduces a mention identification-enhanced EE model, which specializes in clinical mention identification by performing contrastive learning to distinguish correct mentions from the ones with perturbed mention boundary, training an auxiliary mention identification module to learn implicit mention properties, and adding explicit mention markers to hint mention boundaries; 2) performs independent queries for each argument role to better handle long-tail argument roles. 
% We train auxiliary mention identification (MI) module using contrastive learning to improve boundary recognition for all potential triggers, arguments, and entities (hereafter, this collection will be referred to as mentions). The MI module provides explicit mention annotations that are incorporated into the task-specific prompts provided to the generative model. We also perform independent queries for each argument role type to better handle the number of argument roles.
% Even though the techniques are inspired by the traits of the clinical domain events, we show later in this work that they can generalize to other domains that are abundant in lengthy technical terms and long-tail event arguments.

To address the training data availability issue, we introduce \dataname, the first clinical event extraction dataset with argument information, which we derive from clinical experts' annotation on PubMed clinical case reports.

% Evaluation results
We benchmark \modelname on \dataname against several recent event extraction models. 
Experiments show that \modelname achieves state-of-the-art clinical event extraction results on \dataname, and we observe a larger performance gain under low-resource settings. 
Moreover, \modelname also achieves better performance on the ACE05 dataset, demonstrating its generalizability to other domains.
% We also perform ablation studies to demonstrate the effectiveness of each input segment and the design choice of \modelname.

% Contribution summary
Our contributions are threefold: 1) We develop \modelname, a mention-enhanced clinical event extraction model that better identifies mention boundaries and is scalable to many argument roles;
2) We construct the first clinical event extraction dataset with argument annotations;
3) Our model achieves state-of-the-art performance on clinical and news EE and demonstrates more significant performance gains under low-resource settings.

\section{Related Works}
\label{sec:related_works}
\subsection{General Domain Event Extraction}
% Many prior works approach the event extraction problem as the composition of ED and EAE tasks \cite{ahn-2006-stages}. 
Many prior works formulate EE as token-level classification tasks and trained in an ED-EAE pipeline-style \cite{wadden-etal-2019-entity,yang-etal-2019-exploring,ma-etal-2021-eventplus} or optimized jointly \cite{li-etal-2013-joint,yang-mitchell-2016-joint,lin-etal-2020-joint,nguyen-etal-2022-joint}. 
% Other approaches jointly optimize ED and EAE tasks \cite{li-etal-2013-joint,yang-mitchell-2016-joint,lin-etal-2020-joint,nguyen-etal-2022-joint} while incorporating constraints \cite{han-etal-2020-domain,han-etal-2019-joint}, or including a named-entity recognition task to provide an additional supervision signal \cite{Zhao_Liu_Zhao_Wang_2019,ijcai2019-753,sun-etal-2020-recurrent,wadden-etal-2019-entity}.
%Many prior works formulate event extraction \cite{ahn-2006-stages} as a token-level classification task by training separate trigger and argument extraction models in a pipeline style \cite{wadden-etal-2019-entity,yang-etal-2019-exploring,ma-etal-2021-eventplus}, optimizing two tasks with global features jointly \cite{li-etal-2013-joint,yang-mitchell-2016-joint,lin-etal-2020-joint}, incorporating constrains \cite{han-etal-2020-domain,han-etal-2019-joint}, or use named entity recognition (NER) as additional supervision \cite{Zhao_Liu_Zhao_Wang_2019,ijcai2019-753,sun-etal-2020-recurrent,wadden-etal-2019-entity}.
Recent work formulates the EE task as text generation with transformer-based pre-trained language models that prompt the generative model to fill in synthetic \cite{paolini2021structured,huang-etal-2021-document,lu-etal-2021-text2event,li-etal-2021-document} or natural language templates \cite{huang-etal-2022-multilingual-generative,hsu-etal-2022-degree,ma-etal-2022-prompt,ye2022ontology}. 
%\citet{paolini2021structured} uses augmented natural languages to encoder structured information ans formulate IE as machine translation, \citet{huang-etal-2021-document} fills in non-natural language template for document-level EE to capture cross-entity dependencies, and 
% \citet{hsu-etal-2022-degree} jointly optimize the trigger and argument extraction task by generating a human-written template filled in with trigger and argument values, which shows great performance especially on low-resource setting.
These generative EE models are not optimized to handle complicated domain-specific mentions.
To our knowledge, there is no existing approach to clinical EE using a text generation formulation, which we hypothesize is due to both data unavailability and to the aforementioned domain challenges.\looseness=-1

\subsection{Event Extraction in Biomedical Domain}
Biomedical EE is a type of biomedical IE tasks \cite{10.1093/jamia/ocx132,FU2020103526,xu-etal-2023-can-nli}.
Existing approaches to biomedical EE \cite{huang-etal-2020-biomedical,trieu2020deepeventmine,wadden-etal-2019-entity,ramponi-etal-2020-biomedical,wang-etal-2020-biomedical} typically focus on extracting interactions or relationships between biological components such as proteins, genes, drugs, diseases and outcomes related to these interactions \cite{ananiadou2010event}. The mentions in these biological component interactions are short, distinctive biomedical terms and do not have rich event type-argument role ontologies because of the lack of interaction types present in the datasets \cite{ohta-etal-2011-overview, kim-etal-2011-overview-genia, kim-etal-2013-genia, pyysalo-etal-2011-overview, 10.1093/bioinformatics/bts407}. 
\citet{li2020system} develop a clinical event extraction model, but it only handles single-word events without considering arguments \cite{bethard-etal-2016-semeval}.
Our work addresses these concerns by introducing \dataname as well as providing a benchmark in a previously under-explored domain.
%Existing resources include 
% Cancer Genetics 2013, 
%EPI 2011 on protein and DNA event \cite{ohta-etal-2011-overview},
%GENIA on bio-molecular event \cite{kim-etal-2011-overview-genia, kim-etal-2013-genia},
%Infectious Diseases 2011 on bio-molecular and disease interaction \cite{pyysalo-etal-2011-overview},
%Pathway Curation 2013 on molecular event of cancer \cite{pyysalo2015overview},
%Multi-level event extraction on internal event of biological organization from subcelular to the organism level \cite{10.1093/bioinformatics/bts407}, etc.
%These resources and associated models \cite{huang-etal-2020-biomedical,trieu2020deepeventmine,wadden-etal-2019-entity,ramponi-etal-2020-biomedical,wang-etal-2020-biomedical} focus on biological component interactions, in which the event mentions (mostly short and distinctive biomedical terms) are easier to detect, the diversity of event types and argument roles are limited due to the limited physical interaction. 
% extraction in the clinical notes are under-explored due to the lack of resources.

\section{Clinical Domain Event Extraction}
\label{sec:dataset} 

% \violet{I'll probably frame this as ``Event extraction in the clinical domain'', to more generally introduce the domain challenge, and introduce the dataset as a side, since it's no longer our major contribution.}
% \violet{we should probably add a transition paragraph here to talk about the lack of dataset issue for the clinical domain.}
% We first introduce the problem formulation, then provide details about the data composition process and data statistics. %contribute a clinical EE benchmark \dataname to fill in the blank.

% In this section, we introduce the clinical EE task and our novel EE dataset, \dataname, to address the challenges of annotated EE data availability in the clinical domain.

\subsection{Task Formulation}
%\violet{We can actually move this to the beginning of the previous section}
\label{sec:task_formulation}
We follow the framework of prior works that decomposes the EE task into Event Detection (ED) and Event Argument Extraction (EAE), while introducing our novel Mention Identification module as an auxiliary task performed alongside both the ED and EAE modules. ED subtask takes a sentence \texttt{(passage)} as input to extract event triggers and predict event types. The trigger must be a sub-sequence of the \texttt{passage} and the event type must be one of the $n_{event\_type}$ pre-defined types. The EAE subtask takes a tuple of \texttt{(passage, event trigger, event type)}, and extracts arguments from \texttt{passage} and predicts the argument role. Each event type holds a pool of $n_{arg\_role}^{event\_type}$ argument roles as defined in the event ontology.\looseness=-1 
\subsection{The \dataname Dataset}
% \violet{I think it's unclear why/whether the \sourcedata dataset does not have event annotation, and why it's easy to convert \sourcedata into \dataname. I think this part will benefit from some gentle introduction of the \sourcedata dataset}
Due to high annotation costs and privacy concerns, dataset availability is a primary bottleneck for clinical EE. We propose a repurposing of an existing expert-annotated dataset, \sourcedata \cite{maccrobat},\footnote{We use the 2020 version of \sourcedata. We show more details about \sourcedata in \appref{app:maccrobat}.} to compose a clinical EE benchmark, \dataname. 

The \sourcedata dataset consists of 200 pairs of \vfinal{English} clinical case reports from PubMed accompanying annotation files with partial event annotation \vfive{provided by 6 annotators with prior experience in biomedical annotations}. To our knowledge, this is the only openly accessible collection of clinical case reports annotated for entities and relations by human experts. 
% MACCROBAT contains 12 relation types, but for our purposes we only consider the MODIFY relation that occurs when an entity describes or characterizes an event.
Following existing sentence-level EE works \cite{lin-etal-2020-joint}, we construct an event extraction dataset with full event structure, \dataname, which contains annotated span information for \textit{entities}, \textit{event triggers}, \textit{event types}, \textit{event arguments} and \textit{argument roles} for each sentence.
% \mypar{Extracting entities, event triggers and types.} 
% \sviolet{do we need to mention this? To make the construction of \dataname clearer, we can maybe have paragraph titles of ``event trigger construction'', and ``event arguments and roles construction.''}
Mentions are defined as meaningful text spans of occurrences and their properties \cite{maccrobat}. We include all tagged mentions in \sourcedata as \textit{entities}, and further specify that mentions tagged as events and their respective types are included as \textit{event triggers} and \textit{event types}. 
% The \sourcedata dataset annotations assign a single generic tag to all mentions (\ie entities, events, and arguments) and provide an additional annotation tag to specify that a given mention also functions as an event. We categorize all tagged mentions as entities in \dataname to incorporate all available information.

% \mypar{Inferring event arguments and roles.}
\vfive{To infer event arguments and their roles, which are not provided in \sourcedata, we consider non-event entities that hold a \textsc{Modify} relation with event triggers as arguments, and we use the assigned entity types as argument roles.}
% \sihung{You probably need to clarify the entity type here is slightly different from what we usually mean as entity type in general domain NLP?? Otherwise, I would be very confused on why entity types can be serve as argumnent role??}
We infer arguments via the \textsc{Modify} relation because its definition of an entity modifying an event matches well with the argument definition of further characterizing the properties of an event \vfive{as shown in \appref{app:infer_event_arguments}}.
The entity type in \sourcedata defines a type of fine-grained physical or procedure property, which matches the argument role definition of being a type of participant or attribute of an event.
% The \textsc{Modify} relation in the MACCROBAT dataset connects 2 arguments, and defines a "generic relationship in which one entity or event modifies another entity or event, including instances where an entity is identified following an event" \cite{maccrobat}. 
% \begin{figure}
%     \centering
%     \includegraphics[width=\columnwidth]{fig/ev_modify.png}.
%     \caption{MODIFY Relation.}
%     \label{fig:ed}
% \end{figure}
% Thus, given an event trigger, we consider entities that modify the event trigger as arguments of this event. We take the assigned type of the selected entity according to MACCROBAT as the role of the argument. 
We traverse all \texttt{(event type, argument role)} pairs to obtain the argument roles possible for each event type to create an event ontology\vfive{, as shown in \appref{app:event_ontology}}. 
The definitions of each event type and argument role written by clinical experts are provided.
% Thus, in the construction of \dataname, we defined a \textsc{Modify} relation as an event trigger-argument pair in the case where the arguments consist of an entity and an event and the entity modifies the event. For example, Figure 3 illustrates an example of an entity modifying an event that would be classified as an event trigger-argument pair. 
% As previously mentioned, each entity is assigned a type according to ACROBAT, which allowed for the construction of an event type-argument role ontology by observing all event type-entity type pairs. %\derek{Describe how we get the argument role ontology (like for a certain event type, what are possible argument roles, how we get the list of argument roles for each type)}

% More: discontinuous mentions; nested mentions; what information is included in annotation (e.g. entity and type, event and type, relation)

\begin{figure*}[t]
    \centering
    \includegraphics[width=\textwidth]{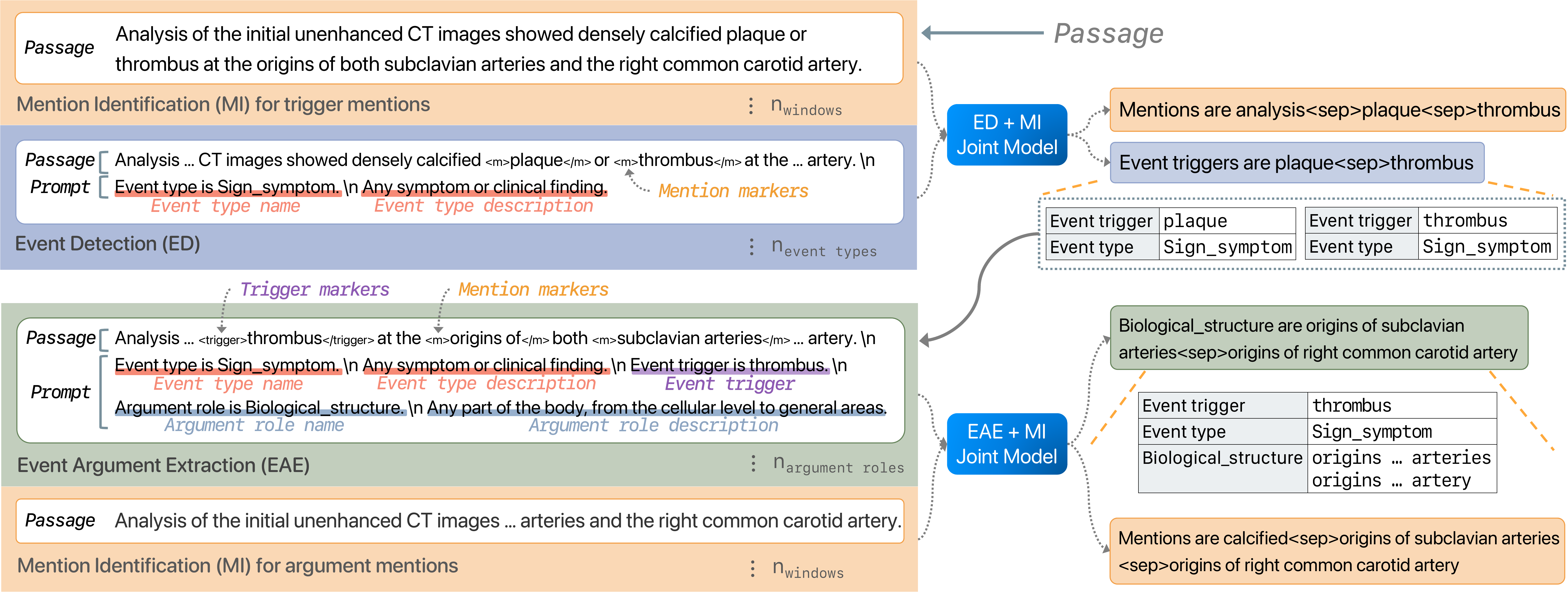}
    \caption{Model design of \modelname. \vfive{We use T5-large \cite{JMLR:v21:20-074} as the backbone text generation model for the two joint models.} 
    % We use the mention identified by the MI module to add mention markers used for the ED/EAE modules. 
    The ED module extracts event trigger and type, and the EAE module extracts argument and roles. They are trained jointly with the trigger and argument MI modules for mention-enhanced event extraction.
    }
    \label{fig:model}
\end{figure*}

\subsection{Data Statistics}
\begin{table}[ht]
\begin{center}
\resizebox{\linewidth}{!}{
{
\small
\setlength\tabcolsep{1.2pt}
\begin{tabular}{lrrr}
\toprule
Metric & ACE05 & ERE & \dataname
\\
\midrule
Unique event types &33 &38 &13 \\
Unique argument roles &22 &21 &22 \\
\vtwo{Unique arg. roles per event type} &4.73 &2.87 &10\\
Documents \# &599 &459 &200 \\
Sentences \# &20,862 &17,114 &4,539 \\
Entities \# &54,820 &46,185 &23,898 \\
Trigger \vtwo{mentions} \# &5,348 &7,287 &13,128 \\
Argument \vtwo{mentions} \# &8,102 &10,479 &8,599 \\
Avg entities \# per sentence &3.18 &3.20 &5.43 \\
Avg events \# per sentence &1.34 &1.47 &3.21 \\
Avg args \# per sentence &2.39 &2.24 &2.67 \\
Avg args per event \# &1.48 &1.42 &0.81 \\
Avg entity word count &1.12 &1.10 &1.89 \\
Avg trigger word count &1.05 &1.06 &1.61 \\
Avg argument word count &1.14 &1.14 &1.72 \\

\bottomrule
\end{tabular}
}
}
\caption{
\vfinal{Statistics of \dataname.}
}
\label{table:data_stats}
\vspace{-0.5em}
\end{center}
\end{table}
In \tbref{table:data_stats}, we show the statistics for \dataname as well as the comparable values for two widely-used EE datasets, ACE05 \cite{doddington-etal-2004-automatic} and ERE-EN \cite{song-etal-2015-light}.
% \appref{app:more_data_stats} shows more metrics.
\dataname differs from general-domain EE datasets because it contains fewer sentences and the average occurrences of entities, triggers, and arguments per sentence are significantly higher. Note that the average length of the entities in \dataname is significantly longer. Besides single-span entities, there are also nested and discontinuous entities used as event arguments in \dataname. This demonstrates that \dataname fills a different niche than ACE05 and ERE-EN and provides a valuable benchmark for EE under a clinical setting with high mention density, and allowing for future work to adapt clinical case report domain-specific features.\looseness=-1

\subsection{Human Verification}
We conduct a human annotation to examine the coverage of the induced arguments and the correctness of their roles. 
% Two annotators are asked to annotate whether the induced arguments are comprehensive and whether argument roles are appropriate or not for the same set of 100 randomly sampled events.
Arguments and their roles in 96\% out of 100 randomly sampled events are considered comprehensive and appropriate by both of the two annotators with consensus.

\section{The \modelname Event Extraction Model}
\label{sec:method}
\vspace{-0.3em}
We formulate EE as a conditional generation task, so that we can incorporate domain knowledge such as event type and argument role definitions via natural language in the input prompt. 
% We first introduce the background of seq2seq extraction models and components of \vfour{Vanilla \modelname, our base model design in \secref{sec:components}}. Then, we introduce our techniques to better leverage mention information \vfour{on top of Vanilla \modelname} in \secref{sec:mention}. Finally, we introduce how to combine the components during training and inference time in \secref{sec:train-inference}.
% Recall our novelty
To tackle the challenges of clinical EE, we 1) further enhance the EE model's specialization in mention identification by techniques introduced in \secref{sec:mention} to handle long clinical mentions with vague boundaries; and 2) perform an independent query for each event type/argument role for better long-tail performance in settings with many event types/argument roles as introduced in \secref{sec:components}.
\figref{fig:model} shows the model design.\looseness=-1

\subsection{Seq2seq Components}
\label{sec:components}
\vspace{-0.3em}
There are three components: 1) Mention Identification (MI) which identifies the candidate pool of event triggers or event arguments, 2) Event Detection (ED) which extracts event triggers and predicts event types, and 3) Event Argument Extraction (EAE) which extracts arguments and predicts argument roles. We integrate these components to form the MI-ED-EAE pipeline (details in \secref{sec:train-inference}). 
\vfive{We use pre-trained text generation model T5-large \cite{JMLR:v21:20-074} as the backbone LM.
The input is a natural language sequence consisting of the original input \textit{passage} and \textit{prompt}. We design input-output formats with shared common elements across different tasks to enable synergistic joint optimization, as all three modules aim to generate a sub-sequence of the input passage.
}

\mypar{Mention Identification (MI).} 
\vspace{-0.3em}
To better align the MI task with the ED and EAE tasks, the MI module extracts all mentions that are candidate event triggers or arguments from the input passage. The input is the \textit{passage} 
and the output includes all trigger or argument candidates in the input passage separated by a special token ``\textsc{[sep]}'' following the prefix ``Mentions are''. If there are no mentions, a placeholder is generated (\ie ``Mentions are $<$mention$>$'').
\vfive{
We extract mentions by inputting the entire passage as well as sentence segments selected by a sliding window with a size of a few words, which enables shorter outputs and higher mention coverage. 
}
% \violet{not clear why we need this. ALso, this (and the following three sentences) feels like implementation details to me. We should probably stay more "conceptual" in the model section}
% The sliding window scan the passage from beginning to end with pre-defined window size and step size, which significantly boosts the coverage of the predicted mentions. During both training and inference, we retain the original full-length input passage in addition to the sliding window segments. 
We enforce the condition that the order of output mentions match the order of their appearance in the passage. This consistency helps the generative model to learn its expected behavior as well as allows for prior mention predictions to inform subsequent mention predictions. We keep the full passages in addition to the sliced sub-sequence during both training and inference to ensure the longer dependencies are captured.
% To distinguish the MI task with ED or EAE during multi-task learning, which will be further discussed in \secref{sec:mention}, the prefix ``Mentions are '' is prepended to the expected output sequence.

\mypar{Event Detection (ED).}
\label{sec:ed}
\vspace{-0.2em}
The ED module extracts event triggers from the passage. 
For a given passage, we construct $n_{event\_type}$ queries. For each query, we input the concatenation of \textit{passage} and the following \textit{prompt} segments:
\textit{event type name} and \textit{event type description}.
% \vspace{-0.3em}
% \begin{itemize}[leftmargin=*]
% \itemsep-0.2em 
%     \item \textit{Event type name}: the name of the query event type such as ``Sign\_symptom'' with a prefix ``Event type is''
%     \item \textit{Event type description}: a brief definition of the event type, such as ``Any symptom or clinical finding''
% \end{itemize}
% \vspace{-0.3em}
The output of the ED task is the concatenation of the event trigger texts predicted for the queried event type separated by a special token ``\textsc{[sep]}'', following the prefix ``Event triggers are''. When there is no valid trigger for the queried event type (which are considered to be negative samples), a special placeholder is generated (\ie ``Event triggers are $<$trigger$>$''). The balance between positive and negative samples is a hyperparameter that may be tuned for a better precision-recall trade-off. 
% To obtain the final predictions for a given passage, 
We decode the output sequence and obtain a list of \texttt{(event type, trigger)} pairs. 
% We then evaluate the correctness of the predicted triggers and their respective types. We consider an event trigger to be correctly identified if it is extracted by any query and to be correctly classified if the queried event type matches the ground truth event type. Compared with existing works that do $n_{event\_type}$ to 1 classification, our design retains the flexibility to capture that triggers can represent multiple events with different types.
% Independent query, no way to see other event type's prediction, can predict multiple types for a trigger
 
\mypar{Event Argument Extraction (EAE).}
\label{sec:eae}
% The input to the EAE model is a event and the sentence containing this event, and the event includes event trigger and event type. The expected output is all arguments and their roles contained in the given sentence that is associated with the given event. Note that even in the same sentence, the arguments for different event triggers could be (and mostly likely) different.
The EAE module extracts event arguments from queries consisting of the input passage, a given role type, and a pair consisting of an event trigger and its event type. We perform $n_{arg\_roles}^{event\_type}$ queries to extract arguments corresponding to each potential argument role where $n_{arg\_roles}^{event\_type}$ is the number of unique argument roles for a certain event type.
%For example, the possible argument roles for the \textsc{Clinical event} are ``Nonbiological location'', ``Detailed description'', ``Frequency'', ``Biological structure'', ``Subject'', ``Lab value'', ``Quantitative concept'', ``Volume'. 
The input sequence contains \textit{passage}, \textit{event type name}, and \textit{event type description} segments in addition to:
\begin{itemize}[leftmargin=*]
\itemsep-0.4em 
    \item \textit{Trigger markers} which are special tokens (\ie ``$<$trigger$>$'' and ``$<$/trigger$>$'') to wrap trigger text to explicitly provide the trigger position
    \item \textit{Trigger phrase} such as ``Event trigger is plaque''
    \item \textit{Argument role name} for the queried argument role, such as ``Argument role is Severity''
    \item \textit{Argument role description}
\end{itemize}
The expected output begins with a reiteration \cite{ma-etal-2023-parameter} of the querying argument role (\eg ``Severity is'') followed by the concatenated predicted argument texts or a placeholder (``$<$argument$>$'') if there are no valid predictions. 
% We then decode and evaluate the generated output.
% according to the same rules outlined in ED.
% Similarly to ED, we can decode the generated output such as ``Event argument is 37 degrees'' and produce the list of predictions consisting of (argument role, argument text) pair such as (\textsc{Lab value}, ``37 degrees'').

\subsection{Mention Identification Enhanced EE} 
\label{sec:mention}
\vfive{
We propose techniques to enhance the generative model's ability to accurately identify long mentions with vague boundaries: \vfour{1) contrastive learning with instances of perturbed mention boundaries, 2) explicit boundary hints with markers and 3) implicit joint mention representation learning}. \looseness=-1
}

\vfive{
\mypar{Contrastive learning with mention boundary perturbation.}
% motivation
Understanding the role of mention descriptors and distinguishing the subtle boundary difference are not specifically optimized during pre-training or fine-tuning with the text generation objective. We propose to create such a task and train the model specifically to recognize the mention with the correct boundaries from a pool of mentions with similar but shifted boundaries.

% creation of negative samples
Following the seq2seq formulation introduced in \secref{sec:components}, we construct $N$ input-output sequence pairs $\langle in_i, out_i \rangle$ where the input sequence $in_i$ consists of passage and prompt, and the gold output $out_i$ contains the ground-truth mentions, triggers or arguments for MI, ED or EAE respectively. 
For a certain input $in_i$, we consider the ground-truth output $out_i$ as a positive output (\eg ``Mentions are ... right common carotid artery''). We create the $k$ negative instances (\ie $n_i^1, ..., n_i^k$) of $in_i$ by perturbing the left and right boundaries of mentions in $out_i$ to add/remove words (\eg removing ``right'', removing ``artery'', or adding ``the'' before ``right'' etc.). We create the negative instances by perturbing output sequences without changing the input, and the contrastive learning objective applies to MI, ED and EAE. This results in a group of instances for $in_i$ including both positive and negative instances:
$\mathbf{X}_{i}=\left\{\left\langle out_i, in_i\right\rangle,\left\langle n_{i}^{1}, in_i\right\rangle, \ldots,\left\langle n_{i}^{k}, in_i\right\rangle\right\}$.
Applying the process, we obtain instance groups for all input-output pairs $\mathbb{X}=\left\{\mathbf{X}_{1}, \ldots, \mathbf{X}_{N}\right\}$.

% loss
We use cross-entropy loss $\mathcal{L}_{CE}$ to learn to generate the correct output $out_i$ given input $in_i$. 
We introduce an InfoNCE loss \cite{oord2018representation} to learn to identify the positive output (items in the numerator) from a pool of output candidates with mention boundary perturbations (items in the denominator) \cite{ma-etal-2021-hyperexpan-taxonomy,meng2021coco,shen2020simple}:\looseness=-1
\begin{equation*}
\mathcal{L}_{N}=\frac{1}{|\mathbb{X}|} \sum_{\mathbf{X}_{i} \in \mathbb{X}}\left[\log \frac{f\left(out_{i}, in_{i}\right)}{\sum_{\left\langle n_{i}^{j}, in_{i}\right\rangle \in \mathbf{X}_{i}} f\left(n_{i}^{j}, in_{i}\right)}\right]
\end{equation*}
where $j \in [0,1,2,...,k]$ and $n_{i}^{0}$ is the positive output $out_i$. 
We define the function $f\left(s, in_i\right)$ as the probability of generating a sequence $s$ given input $in_i$, which is estimated by multiplying logits for each token of the output produced by the decoder under the teacher-forcing paradigm while $in_i$ is fed to the encoder. This estimation is normalized by the output length and produces the output value of $f\left(s, in_i\right)$.
We combine the two losses into the final objective $\mathcal{L}(\Theta)=\mathcal{L}_{CE} + \mathcal{L}_{N}$.\looseness=-1}

\mypar{Explicit mention marker.} 
% \sviolet{I suggest us not citing ourselves in this part to avoid possible identity reveal. It also makes it sounds less novel. We should directly motivate that we want more accurate mention boundary detection and we introduce this explicit marker to help.} Derek: Addressed!
Wrapping key spans with special token markers provides beneficial hints to the generative model that improve its understanding of how the components of the sentence are associated syntactically. 
% In this work, 
We wrap trigger or argument mentions for the ED and EAE tasks, respectively, to provide a candidate pool for the identification task. To minimize the impact of error propagation of the imperfect MI module on downstream tasks, we consider two conditions: 
% \sviolet{the two conditions are a little hard for me to understand. I suggest we create a paragraph for each of the two considerations and make the paragraph title informative so people understand your goal. E.g., ``Robust mention markers'' and ``More precise candidate pool'', something like this...} 
1) the ED/EAE modules with markers must be robust enough to handle the compromised precision and incomplete coverage of the gold mentions and 2) the granularity of the candidate pool must not be too coarse or too fine.
To address the first concern, we generate two versions of the data: one with mention markers and one with no markers, and train the ED/EAE module over the augmented data.
% \footnote{We do not train with marker produced by the standalone MI module as it could overfit low-resource training data.} 
This trains the model to be robust in cases where the MI module provides imprecise predictions.
The second concern stems from the too broad a candidate pool making the markers less informative and too strict a candidate pool making it difficult for the MI module to correctly identify mentions. To account for this issue, we use trigger mentions for the ED task and argument mentions for the EAE task as candidate pools as opposed to using words of a certain part-of-speech or named entities type. The unique properties of triggers (describing an entire process or behavior that can be linked to a specific time) and arguments (concrete details or descriptive content) make them more useful as candidate sets.\looseness=-1

\begin{table*}[t!]
\begin{center}
\small
\resizebox{\linewidth}{!}{
\setlength\tabcolsep{4pt}
\begin{tabular}{cl|
ccc|ccc|
ccc|ccc
}
\toprule
\multirow{3}{*}[-4pt]{\#} &
\multirow{3}{*}[-4pt]{Model} & 
\multicolumn{6}{c|}{Trigger} & \multicolumn{6}{c}{Argument}
\\ \cmidrule{3-14}
& & \multicolumn{3}{c|}{Identification} & \multicolumn{3}{c|}{Classification} & \multicolumn{3}{c|}{Identification} & \multicolumn{3}{c}{Classification}
\\
& & Prec. & Recall & F1 & Prec. & Recall & F1 & Prec. & Recall & F1 & Prec. & Recall & F1
\\
\midrule
1 & Text2Event
& -- & -- & -- & 66.64 & 60.57 & 63.46
& -- & -- & -- & \textbf{55.29} & 47.89 & 51.33
\\
2 & OneIE
& \textbf{74.60} & 74.93 & 74.77 & \textbf{68.74} & 68.96 & 68.85
& 48.99 & 52.59 & 50.72 & 39.82 & 42.95 & 41.32
\\ 
3 & DEGREE
& 71.91 & 66.33 & 69.01 & 67.59 & 62.59 & 65.00 & 46.84 & 24.31 & 32.02 & 44.75 & 23.23 & 30.58

\\ \midrule
4 & Vanilla \modelname 
& 65.03 & 74.08 & 69.26 & 60.51 & 70.28 & 65.03 \vanillaDiceEAEGold
\\
5 & \vfive{\modelname}
& 73.53 & \textbf{76.98} & \textbf{75.22} & 68.12 & \textbf{72.97} & \textbf{70.46} & \textbf{55.41} & \textbf{57.87} & \textbf{56.61} & 53.02 & \textbf{55.03} & \textbf{54.01}
\\
\bottomrule
\end{tabular}
}
\caption{
Event detection and event argument extraction performance (\%). The EAE task takes the predicted event trigger and event type as input from the corresponding ED model in the pipeline style. \modelname achieves the state-of-the-art event trigger and argument identification and classification performance.
}
\label{table:overall}
\end{center}
\end{table*}
\mypar{MI as an implicit auxiliary task.} 
\vfive{Existing works include a named-entity recognition task to provide additional supervision signals for EE \cite{Zhao_Liu_Zhao_Wang_2019,ijcai2019-753,sun-etal-2020-recurrent,wadden-etal-2019-entity} for other formulations except for generative models.}
Since we design all three extraction tasks (ED, EAE and MI) as generation tasks, % that extract a sub-sequence from the input passage given a task-specific prompt. 
% Old motivation of using aux task
% Compared to sequence tagging models with complex decoding designs, generative models are not as adept at extracting long and multi-token concepts despite the advantages we introduced in the beginning of \secref{sec:components},\sviolet{where? I couldn't find...} especially in a domain it is not pre-trained for. \sviolet{I don't think we need to motivate it this way. I feel we can directly say we do multitask learning, which is an intuitive way to jointly train multiple interdependent tasks and enhance performance.} 
% New motivation of using aux task 1101
% Given identifying meaning spans is a synergistic capability contributing to performing ED and EAE, 
and ED and EAE can be considered as special MI with certain interest focus, identifying mentions is a synergistic capability contributing to performing ED and EAE.
Thus, we add trigger MI and argument MI as auxiliary tasks to jointly optimize with the ED and EAE tasks, respectively.
\looseness=-1

\subsection{Training and Inference}
\label{sec:train-inference}
\mypar{Schedule sampling.} To gently bridge the discrepancy between gold and predicted upstream results (ED results passed to EAE, trigger/argument MI results passed to ED/EAE), we adopt the scheduled sampling technique to perform curriculum learning \cite{bengio2015scheduled}. We force the downstream model to deal with imperfect upstream results gradually by decaying the upstream results from the gold ones to the predicted ones linearly. We perform the decay at the beginning of each epoch.

\mypar{Training.} 
% We provide ground-truth upstream data points (\ie mention markers for ED and EAE, event triggers and types for EAE) when separately training the ED, EAE, and MI modules and use the standard teacher forcing cross-entropy text generation loss to train the model.
We first train standalone trigger and argument MI modules to provide mention candidates. We then train ED+MI joint model and EAE+MI joint model with auxiliary trigger and argument MI modules respectively. We also add markers around trigger/argument mention candidates.
% Finally we train an EAE module with auxiliary argument MI module with ground-truth mention markers, event trigger and event types.
For efficient training, the model uses downsampled negative instances (\ie instances with mismatched trigger/argument and event type/argument role).
% for faster convergence.

\mypar{Inference.} We use the trigger and argument mention markers produced by the standalone trigger and argument MI modules in the downstream ED+MI and EAE+MI joint models. The event triggers and their types predicted by the ED+MI joint model are provided as input to the EAE+MI joint model in a pipeline fashion.

\section{Experiments in the Clinical Domain}
\label{sec:eval}
We evaluate \modelname on \dataname and compare it with existing event extraction models.

\subsection{Experimental Setup}
\label{sec:experimental_setup}

\mypar{Data splits.}
We divide the 200 \dataname documents according to an 80\%/10\%/10\% split for the training, validation, and testing sets, respectively.
% For each data split, we use the individual sentences and annotated mentions in corresponding documents as data instances for the ED and EAE tasks.
For low-resource settings, we consider 10\%, 25\%, 50\%, and 75\% of the number of \textit{documents} used to build the training dataset while retaining the original validation and testing sets for evaluation.

\mypar{Evaluation metrics.} We follow previous EE works and report precision, recall and F1 scores for the following four tasks.
    % 1) Trigger Identification: a trigger is correctly identified if it matches the ground truth span.
    % 2) Trigger Classification: a trigger is classified if it is correctly identified and its predicted \textit{event type} matches the ground truth event type.
    % 3) Argument Identification: an argument of an event is correctly identified if it matches the ground truth span.
    % 4) Argument Classification: an argument of an event is classified if it is correctly identified and its predicted \textit{argument role} matches the ground truth argument role.
    1) Trigger Identification: identified trigger span is correct.
    2) Trigger Classification: identified trigger is correct and its predicted \textit{event type} is correct.
    3) Argument Identification: identified argument span is correct.
    4) Argument Classification: identified event argument is correct and its predicted \textit{argument role} is also correct.
%Note that despite entity annotations being provided in \dataname, they are incorporated into our model as our design has a better performance shown in \secref{sec:ablation_mention}.

\vfive{
\mypar{Variants.} 
We term two variants of our model. 
We refer to pipelined ED and EAE modules \textit{without} the mention enhancement techniques described in \secref{sec:mention}, \textit{with} long-tail argument handling and text generation cross-entropy loss only as \textbf{Vanilla \modelname}, and the full model as \textbf{\modelname}.}\footnote{We show hyperparameters, implementation and baseline reproduction details in \appref{sec:implementation_details}. 
% We report the median result for five runs by default.
}
% \modelname: pipelined ED and EAE module using mention markers and the mention identification auxiliary task joint training.

\begin{table*}[th]
\resizebox{\linewidth}{!}{
{
\small
\setlength\tabcolsep{3.15pt}
\begin{tabular}{Hcl|
ccc|ccc|
ccc|ccc
}
\toprule
\multirow{3}{*}[-4pt]{\makecell[l]{\vfour{Mention} \\ \vfour{granularity}}} &
\multirow{3}{*}[-4pt]{\#} &
\multirow{3}{*}[-4pt]{\makecell[l]{Mention-enhancing techniques}} & 
\multicolumn{6}{c|}{Trigger} & \multicolumn{6}{c}{Argument}
\\ \cmidrule{4-15}
& & & \multicolumn{3}{c|}{Identification} & \multicolumn{3}{c|}{Classification} & \multicolumn{3}{c|}{Identification} & \multicolumn{3}{c}{Classification}
\\
& & & Prec. & Recall & F1 & Prec. & Recall & F1 & Prec. & Recall & F1 & Prec. & Recall & F1
\\
\midrule
% \cmidrule{2-17}
\textit{\vfour{---}} & 1 & Vanilla \modelname
\vanillaDiceED \vanillaDiceEAE
\\ \midrule
& 2 & Vanilla w/ aux. task
& 69.54 & 74.59 & 71.98 & 65.02 & 71.00 & 67.88
& 73.24 & 76.48	& 74.83	& 68.31	& 73.03	& 70.59
\\
& 3 & Vanilla w/ marker
& 72.91 & 70.71 & 71.79 & 68.58 & 67.70 & 68.14
& 74.27 & 76.91 & 75.57 & 69.66 & 72.82 & 71.20
\\
& 4 & \vfive{Vanilla w/ contrastive}
& 70.02 & 75.12 & 72.48 & 66.93 & 72.04 & 69.39 & 73.86 & 77.41 & 75.59 & 69.92 & 72.89 & 71.37
\\
& 5 & \vfive{Vanilla w/ all three (Full \modelname)}
& 73.53 & 76.98 & \textbf{75.22} & 68.12 & 72.97 & \textbf{70.46} & 75.73 & 77.62 & \textbf{76.66} & 71.14 & 73.91 & \textbf{72.50}
\\ \cmidrule{2-15}
\multirow{-4}{*}[2pt]{\textit{\makecell[l]{\vfour{Fine-} \\ \vfour{grained}}}} & 6 & Vanilla w/ perfect marker\textsuperscript{\dag}
& 97.04 & 94.11 & 95.55 & 85.23 & 88.66 & 86.91
& 91.91 & 90.72 & 91.31 & 81.71 & 86.73 & 84.14
\\
\bottomrule
\end{tabular}
}
}
\caption{
Ablation study of the technique used to incorporate mention information. The argument extraction reported here uses ground-truth event trigger and type, which removes error propagation from the upstream ED result. \dag\ indicates the settings use mention markers to wrap ground-truth mentions and they are not comparable with other lines. \looseness=-1
}
\label{table:ablation_mention-usage}
% \end{center}
\end{table*}

\mypar{Baselines.} We benchmark the performance of the recent EE models on \dataname, including: \textbf{Text2Event} \cite{lu-etal-2021-text2event}: a sequence-to-structure model that converts the input passage to a trie data structure to retrieve event arguments; \textbf{OneIE} \cite{li-etal-2013-joint}: a multi-task EE model trained with global features;\footnote{Note that additional entity annotation is used during training, while it is not used in other models.} and \textbf{DEGREE} \cite{hsu-etal-2022-degree}: a prompt-based generative model that consists of distinct ED and EAE modules that fill in event type-specific human written templates. To adapt DEGREE to the new dataset, we create the ED/EAE templates by concatenating event type/argument role phrases (\eg ``Biological\_structure is artery'').

\subsection{Overall ED and EAE Results}

We show the superiority of \modelname in both high-resource and low-resource settings.

\mypar{High-resource results.} \tbref{table:overall} shows the results for high-resource settings. Among the baselines, OneIE and Text2Event achieve the best F1 score on trigger extraction and argument extraction respectively. DEGREE reports low performance on the argument extraction task due to the challenges of generating long sequences containing all argument roles. \modelname outperforms the baselines on \textit{both} trigger and argument extraction tasks, with 2.7 points F1 score improvements for argument classification.\looseness=-1

\begin{figure}[tbh]
    \centering
    \includegraphics[width=\columnwidth]{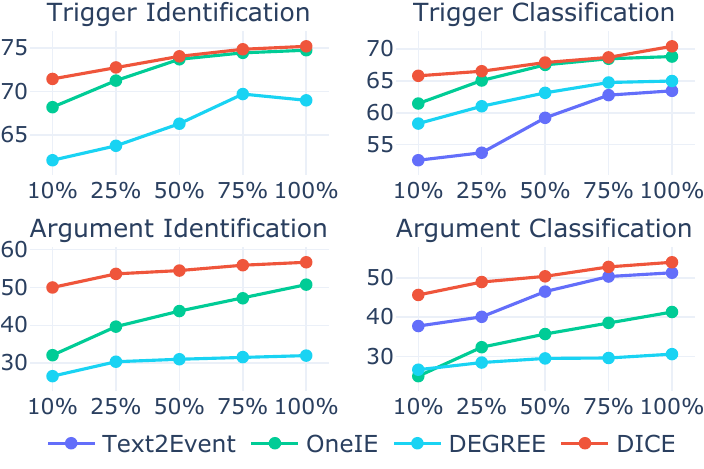}
    \caption{Performance on downsampled training data. 
    We report F1 score (\%, y-axis) for each proportion (x-axis). 
    \vtwo{\modelname outperforms all baselines across four tasks.
    %, and it shows more advantages for the lower-resource settings.
    }
    %We report the averaged F1 score (y-axis) of three random splits for each proportion (x-axis).
    }
    \label{fig:downsample}
\end{figure}

\mypar{Low-resource results.} We show the results of training in lower-resource settings in \figref{fig:downsample} and \appref{app:full_low_resource_results}. 
% Despite OneIE achieving the best performance on ED tasks in higher resource settings, we observe that \modelname outperforms OneIE with a large margin on both trigger extraction tasks in the 10\% data setting and marginally outperforms OneIE on trigger classification in the 25\% data setting.
We observe that \modelname outperforms all baselines on all four tasks under all low-resource settings. The performance gap between \modelname and the baselines increases in the lower training data percentage settings. In the argument classification task, \modelname outperforms Text2Event by more than 8 (10\%) and 9 (25\%) points in F1 score.

\vspace{-1em}
\subsection{Ablation Studies}
\label{sec:ablation_mention}
% \sviolet{I feel this section is less important. we can move it to the appendix if we need space. if we have space, I suggest us to move this as a subsubsection/paragraph in ablation studies.} -> Derek: applied. We removed the comparison between entity/trigger/argument identification, and merged in the ablation studies section. We put the ablation about MI module design after mention-enhancing techniques, mention granularity.
We show ablation studies about mention-enhancing techniques and MI module design in this section and more studies about input prompt segments and formulation in \appref{app:additional_ablation}.

\mypar{Mention-enhancing techniques.}
We analyze the effects of the proposed mention-enhancing techniques in \tbref{table:ablation_mention-usage}.
We observe contrastive learning, auxiliary task, and mention markers contribute improvements of 1.92, 1.14 and 1.75 in the F1 score on argument classification, respectively. We observe that \modelname improves over vanilla \modelname by 5.43 and 3.05 in the F1 score for trigger and argument classification, respectively. 
% This result indicates that the ED and EAE modules benefit from both the implicit joint learning signal gained from the auxiliary task and the explicit mention hints provided by including mention markers in the input sequence. 
We include an oracle setting on Line 6 that provides ground-truth mention markers during inference to illustrate the influence of the accuracy of the MI module.\looseness=-1

\begin{table}[th]
\begin{center}
\resizebox{\linewidth}{!}{
{
\small
\setlength\tabcolsep{2pt}
\begin{tabular}{cl|ccc}
\toprule
\# & Model & Prec. & Recall & F1
\\
\midrule
1 & \citet{yan-etal-2021-unified-generative}
& 72.00 & 72.70 & 72.30
\\
2 & OneIE entity identification module
& \textbf{75.88} & 77.86 & 76.86
\\
3 & \vfive{\modelname-MI without sliding window}
& 71.71 & 67.13 & 69.34
\\
4 & \vfive{\modelname-MI without constrative learning}
& 71.80 & 84.14 & 77.48
\\
5 & \vfive{\modelname-MI}
& 74.20 & \textbf{86.04} & \textbf{79.68}
\\
\bottomrule
\end{tabular}
}
}
\caption{
\vfour{Ablation study of MI module design. 
}
}
\label{table:overall_ET}
\end{center}
\end{table}
\mypar{MI module design.}
We compare our MI module with the representative of sequence tagging model OneIE, which produces BIO label for each input token, and state-of-the-art generative named-entity recognition model \citet{yan-etal-2021-unified-generative}, which generates token indexes, on the entity identification task. We report the performance in \tbref{table:overall_ET}. 
The results show that the sliding window technique significantly improves recall (Line 5 vs 3) and contrastive learning improves overall performance (Line 5 vs 4). Our MI module outperforms all baselines and achieves the best F1 score.

\subsection{Error Analysis}
\begin{figure}[t]
    \centering
    \begin{subfigure}[b]{\columnwidth}
         \centering
        \includegraphics[width=\columnwidth]{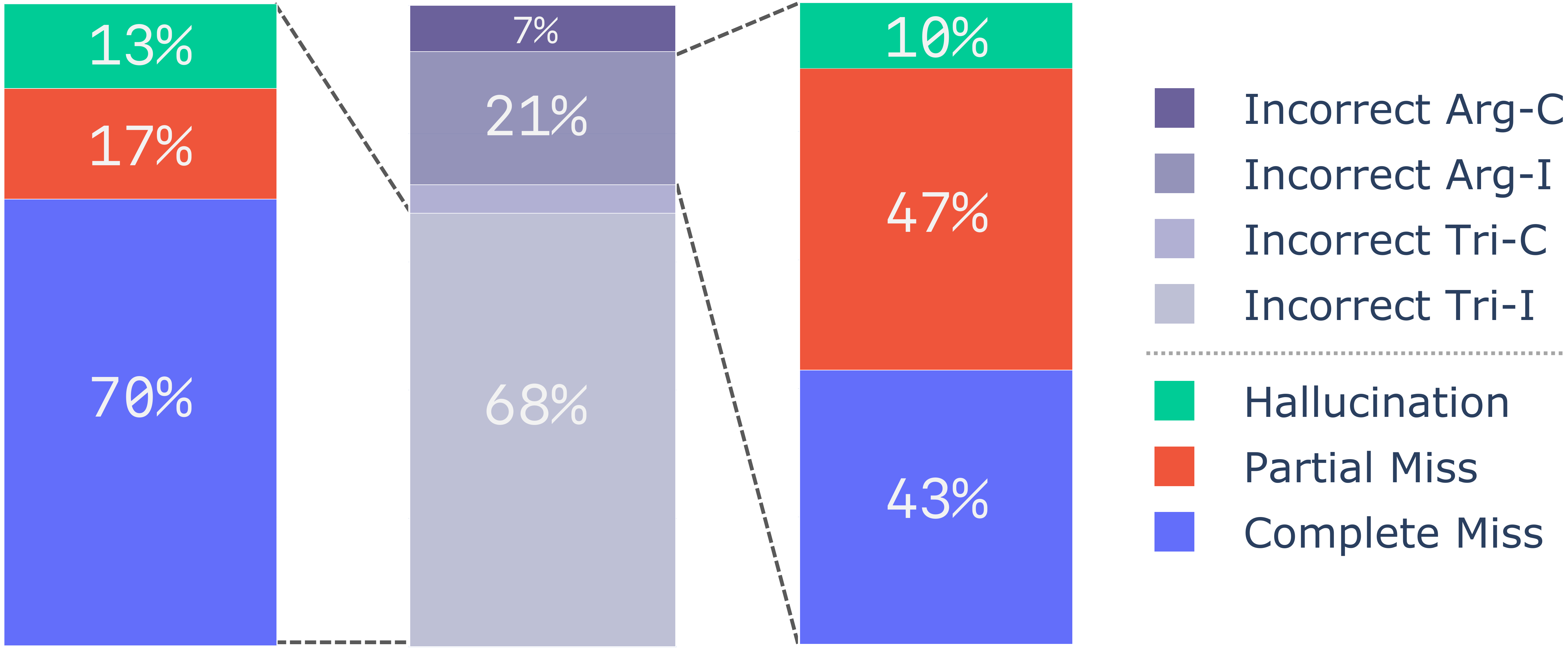}
         \caption{\vfinal{Proportion of error cases by steps in the pipeline.}}
         \label{fig:error_analysis_relative}
     \end{subfigure}
     \begin{subfigure}[b]{\columnwidth}
         \centering
        \includegraphics[width=\columnwidth]{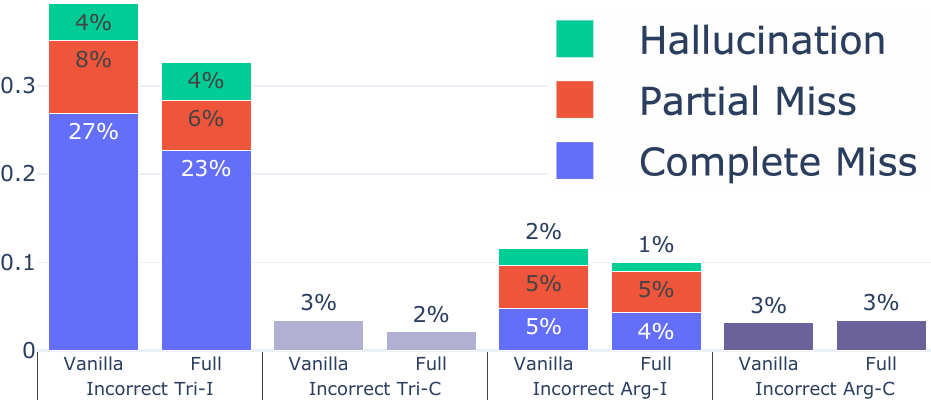} % {fig/error_analysis_comparison.pdf}
         \caption{
         \vtwo{Error produced by each step of the vanilla and full version of \modelname. We show \textit{absolute} error ratios (wrong predictions among all predictions, the lower, the better).}}
         \label{fig:error_analysis_comparison}
     \end{subfigure}
     % \vspace{-1.9em} 
    \caption{
    Error analysis of the argument classification task, which shows the culmination of the errors propagated through the pipeline of \modelname.
    }
    % \vspace{-2em}
    \label{fig:error_analysis}
\end{figure}

We analyze the errors propagated through the 4 steps in the pipeline for \modelname using predicted triggers on the argument classification task which shows the culmination of the errors propagated through the pipeline.
%: 1) The trigger is not correctly identified; 2) The trigger is identified but is classified incorrectly; 3) The ED result is correct, but the argument is not correctly identified; 4) The ED result is correct and the argument is correctly identified, but the argument is classified incorrectly. 
The results in \figref{fig:error_analysis_relative} indicate that the identification sub-tasks, especially trigger identification, are the performance bottlenecks.

We further break identification errors into three types: 1) complete miss: the predicted span has no overlap with the ground-truth span; 2) partial miss: the predicted span is a subset of the ground-truth span; 3) hallucination: the predicted span partially overlaps with the ground-truth span, but also incorrectly includes additional tokens. 
As shown in \figref{fig:error_analysis_relative}, the majority of errors produced by the trigger identification step are complete misses, whereas argument identification suffers from both partial and complete misses. We also observe that the left boundaries of the trigger and argument spans are more difficult to identify as 76\% of partial misses and 69\% of hallucinations correctly identify the right boundary but miss the left boundary. This can be explained by that the dominant word of the entity is typically on the rightmost (\eg ``attack'' in ``heart attack''), whereas the left boundary requires separating the target entity from its descriptors (\eg ``massive heart attack'').

We further compare the error types between the vanilla \modelname and full version of \modelname with mention identification enhancement techniques in \figref{fig:error_analysis_comparison}. We observe that \modelname produces fewer error cases across all error types in both trigger and argument identification steps, which supports our assertion that our mention identification enhancement techniques improve the identification of mentions with vague boundaries.

\newcommand{\trigger}[1]{
\textcolor{orange}{#1}
}

\newcommand{\gt}[1]{
\hlc{lightblue}{#1}
}

\begin{table*}[h]
\begin{center}
\resizebox{\linewidth}{!}{
{
\footnotesize
\setlength\tabcolsep{4pt}
\begin{tabular}{ll}
\toprule
\multirow{2}{*}{1}&\textbf{Task}: ED ~~~\textbf{Passage}: An \gt{\{audiology evaluation\}} showed \gt{\{severe\}} \gt{\{bilateral\} \{sensorineural\} \{high-frequency\} \{hearing loss\}} (\gt{\{-70 dB\}}).
\\
&\textbf{Ground-truth}: (bilateral sensorineural high-frequency hearing loss, Sign\_symptom)
~~~
\textbf{Pred. of \modelname}: (hearing loss, Sign\_symptom)
\\ \midrule
\multirow{2}{*}{2}&\textbf{Task}: EAE ~~~\textbf{Passage}: The patient underwent a \gt{\{resection\}} of the \{\gt{15 cm} \gt{segment IVb}\} \trigger{mass [\textsc{Sign\_symptom}]} in \gt{\{June 2010\}}.
\\
&\textbf{Ground-truth}: (15 cm, Distance), (segment IVb, Biological\_structure)
~~~
\textbf{Pred. of \modelname}: (15 cm segment IVb, Biological\_structure)
\\ \midrule
\multirow{2}{*}{3}&\textbf{Task}: ED ~~~\textbf{Passage}: Core \gt{biopsies} from the \gt{\{breast lump\}} showed \gt{\{ductal carcinoma\}} in situ (sample labelled P1.1).
\\
&\textbf{Ground-truth}: (biopsies, Diagnostic\_procedure), (ductal carcinoma, Disease\_disorder)
~~~
\textbf{Pred. of \modelname}: \textit{None}
\\ \midrule
\multirow{3}{*}{4}&\textbf{Task}: EAE ~~~\textbf{Passage}: Serum \gt{total bilirubin} and \gt{\{tumor markers\}},
\trigger{carcinoembryonic antigen [\textsc{Diagnostic\_procedure}]} (\gt{\{CEA\}}) and
\\ & 
\trigger{carbohydrate antigen 19-9 [\textsc{Diagnostic\_procedure}]} (\gt{\{CA19-9\}}), 
were all \gt{\{within \{normal ranges\}\}}.
\\
&\textbf{Ground-truth}: \textit{None}~~~\textbf{Pred. of \modelname}: (within normal ranges, Lab\_value) was predicted as the argument for both events.
\\
\bottomrule
\end{tabular}
}
}
\caption{Qualitative analysis. We mark \textcolor{orange}{event trigger [\textsc{Event\_type}]}, \gt{ground-truth mentions} and \{mention prediction\} made by our MI module.}
\label{table:qualitative_analysis}
\end{center}
\end{table*} 
\vtwo{
\subsection{Qualitative Analysis}
To identify challenges for future works, we summarize 4 types of common errors made by \modelname and show examples in \tbref{table:qualitative_analysis}. 
% \sviolet{please verify whether I'm citing the correct table...}-> Derek: Yes you are! 
In the first example, the MI module of \modelname only identifies a subsequence of the true mention (e.g., ``hearing loss'' vs. ``bilateral sensorineural high-frequency hearing loss''), leading to a partial miss that shows the ED module mistakenly includes incorrect descriptors. 
In the second example, \modelname hallucinates that a \textsc{Distance} descriptor ``15 cm'' is part of the \textsc{Biological\_structure} ``segment IVb'', which indicates that the EAE module struggles to separate mention boundaries. 
In the third example, the first event ``biopsies'' is missed by both the ED module and the MI module. However, despite the MI module correctly identifying ``ductal carcinoma'' as a mention, the ED module does not identify it as an event trigger. 
In the fourth example, \modelname identifies ``within normal ranges'' as the \textsc{Lab\_value} for the two \textsc{Diagnostic\_procedure} events, which are not valid \textsc{Lab\_value} for tumor marker tests. 
% We observe multiple errors introduced by the lack of domain knowledge. For example, some diagnostic procedures assign categorical values to their results (\eg ``high'', ``normal'' or ``low'' for ``mitotic rate''), while others use numeric values (\eg ``219 U/mL'' for carbohydrate antigen 19-9 and other tumor marker tests).

% \mypar{Fail to identify mentions.}

% \mypar{Fail to copy.}

% \mypar{Lack of cross-sentence information.}

% \mypar{Lack of in-domain knowledge.}
}

\section{Experiments in the General Domain}
\label{sec:eval_general_domain}
We evaluate \modelname's generalizability by performing
% by comparing its performance on a general domain event extraction dataset against several baselines. 
% \mypar{Experimental setup.} 
EE on the widely-used news-domain dataset ACE05 \cite{doddington-etal-2004-automatic}, which contains 33 event types and 22 distinct argument roles. 
We perform both full-shot and low-resource experiments with 10\% of the training data using the same data pre-processing, data splits and metrics as prior works \cite{wadden-etal-2019-entity,lin-etal-2020-joint}, and 
we compare with the same set of baselines introduced in \secref{sec:experimental_setup}. Baseline selection criteria and more results are presented in \appref{sec:general_domain_result_additional}.
% \footnote{\vfive{We select published EE works performing ED and EAE tasks \textit{only} without external resources or additional tasks as our baselines.}}

% \mypar{Experimental results.} 
We show the result in \tbref{table:overall_general_domain_simple}. We observe that \modelname achieves a better performance on both low and high-resource settings for both trigger and argument classification tasks. We observe that DEGREE's performance is much closer to our model than in the clinical domain, which is due to two factors. First, the benefit of the independent query design used in \modelname is diminished because ACE05 has far fewer argument roles that need to be filled in for each event type (on average 4.73) compared with in \dataname (on average 10).
Second, DEGREE benefits from the implicit argument role dependencies established in its human-created event templates for ACE05, 
which were unavailable for the clinical domain. 
We also observe that mentions in the general domain are easier to identify as our MI module achieves 92\% F1 score for entity identification on ACE05, while achieving 77\% F1 score on \dataname. 
Although the mentions in the general domain are not as complex as clinical mentions, the performance of \modelname supports our claim that mention-enhanced event extraction generalizes to the general domain.

\begin{table}[t]
\begin{center}
\small
\setlength\tabcolsep{4pt}
\begin{tabular}{Hl|cc|ccHHHH}
\toprule
\multirow{2}{*}[-1pt]{\#} &
\multirow{2}{*}[-1pt]{Model} & \multicolumn{2}{c|}{10\%} & 
\multicolumn{2}{c}{100\%} 
\\
& & \multicolumn{1}{c}{Tri-C} & \multicolumn{1}{c|}{Arg-C} & \multicolumn{1}{c}{Tri-C} & \multicolumn{1}{c}{Arg-C}
\\
\midrule
6 & Text2Event
& 47.0\textsuperscript{\ddag} & 24.9\textsuperscript{\ddag}
& 71.9\textsuperscript{\dag} & 53.8\textsuperscript{\dag}
& 42.4 & 28.7
& \textbf{59.4} & 48.3
\\
7 & OneIE
& 61.5\textsuperscript{\ddag} & 26.8\textsuperscript{\ddag}
& \underline{74.7}\textsuperscript{\dag} & \underline{56.8}\textsuperscript{\dag}
& 48.8 & 35.1
& 57.0 & 46.5
\\ 
8 & DEGREE
& \underline{66.1}\textsuperscript{\dag} & \underline{42.1}\textsuperscript{\dag}
& 72.2\textsuperscript{\dag} & 55.8\textsuperscript{\dag} 
& 50.6 & 39.4
& 57.8 & \textbf{50.4}
\\ \midrule
9 & \vfive{\modelname}
& \textbf{68.9} & \textbf{44.7}
& \textbf{75.5} & \textbf{57.6}
\\
\bottomrule
\end{tabular}
\caption{
\vtwo{
ED and EAE performance (\%) on the general domain dataset ACE05. We report the numbers from the original paper (\dag) or \cite{hsu-etal-2022-degree} (\ddag). 
\textbf{Boldface} denotes the best results while \ul{underscore} denotes the second best.
\modelname achieves state-of-the-art performance across both resource settings and tasks.
}
}
\label{table:overall_general_domain_simple}
\end{center}
\end{table}

% \vspace{-0.6em}
\section{Conclusion and Future Work}
\label{sec:conclusion}
% \vspace{-0.5em}

We present \modelname, a generative event extraction model designed for the clinical domain. 
% Our approach formulates the EE task as a conditional generation task that leverages domain knowledge in the form of a natural language prompt. 
\modelname is adapted to tackle long and complicated mentions by conducting contrastive learning on instances with mention boundary perturbation, jointly optimizing EE tasks with the auxiliary mention identification task as well as the addition of mention boundary markers. 
% \modelname is adapted to tackle long and complicated mentions by conducting contrastive learning, jointly optimizing with the auxiliary mention identification task and adding mention boundary markers. 
We also introduce \dataname, the first clinical EE dataset with argument annotation as a testbench for future clinical EE works. 
Lastly, our evaluation shows that \modelname achieves state-of-the-art EE performance in the clinical and news domains. 
In the future, we aim to 
% pre-train generative language models on a clinical corpus as well as to 
apply transfer learning from higher-resource domains.
% to improve performance.
%to improve its representational power. 
%Event extractions trained on a large amount of news data learn ability event signal from text but lack biomedical domain knowledge. We plan to use the prompt tuning technique \cite{lester-etal-2021-power} that prompts the model with domain information, so that we could use news domain training data to benefit the EE task in the biomedical domain.

\section*{Acknowledgments}
Many thanks to I-Hung Hsu, Derek Xu, Tanmay Parekh and Masoud Monajatipoor for internal reviews, to lab members at PLUS lab, ScAi and UCLA-NLP for suggestions, and to the anonymous reviewers for their feedback. This work was partially supported by NSF 2106859, 2200274, AFOSR MURI grant \#FA9550-22-1-0380, Defense Advanced Research Project Agency (DARPA) grant \#HR00112290103/HR0011260656, and a Cisco Sponsored Research Award. 

\section*{Limitations}
% While we are open to different types of limitations, just mentioning that a set of results have been shown for English only probably does not reflect what we expect. Mentioning that the method works mostly for languages with limited morphology, like English, is a much better alternative. In addition, limitations such as low scalability to long text, the requirement of large GPU resources, or other things that inspire crucial further investigation are welcome.
This work presents a repurposing of an existing dataset, \sourcedata, and a set of novel techniques for adapting event extraction to the clinical domain. Among these new techniques is the handling of long-tailed argument roles, in which we independently query each role type. This presents an issue with scalability to domains with yet more complexity, as training the full \modelname while querying both all event types and all argument types present in \dataname requires considerable resources during inference.  
\section*{Ethical Statement}
% The consideration of the ethical impact of our research, use of data, and potential applications of our work has always been an important consideration, and as artificial intelligence is becoming more mainstream, these issues are increasingly pertinent.

Our experiments and proposed model framework are intended to encourage exploration in the clinical information extraction domain while avoiding the risk of privacy leakage. The data we use in this work is publicly available and fully de-identified. Though recent research has found it to be difficult to reconstruct protected personal information from such data, there remains some small risk that future models may be able to do so. We have not altered the content of data in any that would increase the likelihood of such an occurrence and are thus not risking private information leakage.

\bibliography{anthology_small,custom}
\bibliographystyle{acl_natbib}

\clearpage
\appendix
\section{Potential Questions}
% \subsection{Methods and Implementation}
\mypar{What is the difference between the existing generative EE model DEGREE and \modelname?} Compared with DEGREE, our model: 1) further enhances the EE model's specialization in mention identification by three techniques to learn mention-related capabilities introduced in \secref{sec:mention} to handle long clinical mentions with vague boundaries; and 2) performs an independent query for each argument role for better long-tail performance in settings with many argument roles as introduced in \secref{sec:components}.

% \mypar{How to access the codebase and proposed dataset?} We will make the codebase and the proposed dataset public after acceptance. We also attached the source code and the full dataset in the Softconf submission.

\mypar{Would training and inference efficiency be an issue?} As we perform an independent query for each event type/argument role in the ED/EAE model, it is a tradeoff between performance and running cost. Though during training, we only sample a subset of negative instances to train the model for faster convergence. For example, to create seq2seq input-output pairs for a certain sentence for ED, we create 1 positive pair (\ie there is an event in the sentence for the query event type) and $k$ (instead of $n_{event\_type}$, where $k$ is much smaller than $n_{event\_type}$) negative pairs (\ie no event exists for the query event type).

\mypar{Why use standalone MI modules to produce mention candidates?} We use standalone trigger and argument MI modules to create markers for downstream ED+MI and EAE+MI joint models, instead of using the MI module jointly trained in the ED+MI or EAE+MI models because the standalone one yields better performance.

% \subsection{Experiments}
% \mypar{Why not compare with some works claiming higher performance on ACE?} We select our baselines for both \dataname and ACE05 by criteria stated in \appref{sec:general_domain_result_additional}: we select published EE models reporting performance on the ACE05 dataset using ED and EAE training data only without additional tasks such as relation extraction or entity recognition. Some works (such as \citet{nguyen-etal-2022-learning,nguyen-etal-2021-cross,zhang-ji-2021-abstract}) claim higher F1 scores but they either jointly train with additional tasks other than ED and EAE, or the codebase is not available for us to remove the training signals from other tasks for a fair comparison.
\section{Dataset \dataname Details}
\label{appendix:data}

\subsection{\sourcedata Annotation} 
\label{app:maccrobat}
\sourcedata is annotated according to the Annotation for Case Reports using Open Biomedical Annotation Terms (ACROBAT) defined in \cite{maccrobat}. ACROBAT describes events and entities as meaningful text spans, but differentiates events as occurrences that may be ordered chronologically and entities as objects that may modify or describe events. \vfive{According to the annotation guideline, entity text spans are limited to the shortest viable length.} Each event and entity is given a type such that certain events are associated with certain argument roles. According to ACROBAT, Entity text spans are limited to the shortest viable length. For example, the text span ``mild asthma attack'' would be annotated by labeling ``asthma attack'' as an event as that is the shortest span that conveys the occurrence of the event. ``Mild'' would be labeled an entity and the annotation would add a relation indicating that ``mild'' modifies ``asthma attack''. \sourcedata contains 12 relation types, but for our purposes we only consider the \textsc{Modify} relation that occurs when an entity describes or characterizes an event.
 
\subsection{Details of Inferring Event Arguments}
\label{app:infer_event_arguments}
According to ACE2005 English Events Guidelines (AEEG),\footnote{\url{https://www.ldc.upenn.edu/collaborations/past-projects/ace/annotation-tasks-and-specifications}} the arguments of events are defined as entities and values within the scope of an event and only the closest entities and values will be selected, where a value is defined to be ``a string that further characterizes the properties of some Entity or Event''.
The \textsc{Modify} relation in the \sourcedata dataset connects 2 arguments, and it is defined as the ``generic relationship in which one entity or event modifies another entity or event, including instances where an entity is identified following an event'' \cite{maccrobat}. The \textsc{Modify} relation satisfies the argument definition described by the AEEG by incorporating within-sentence relationships between an entity that modifies or describes an event. Thus, given a certain event trigger, we consider non-event entities that hold a \textsc{Modify} relation with the trigger as arguments of this event. We take the assigned type of the selected entity according to \sourcedata as the role of the argument. To create an event ontology, which includes all possible event types and possible argument roles or each event type, we traverse all (event type, argument role) pairs to obtain the unique argument roles possible for each event type.

\subsection{Event Ontology}
\label{app:event_ontology}
We show the full event ontology, including all event types and their possible argument roles, in \tbref{table:event_ontology}.

% \subsection{More Data Statistics}
% \label{app:more_data_stats}
% \input{table/data_stats}
% We presented more detailed data statistics including more metrics in \tbref{table:data_stats_appendix}.
\section{Additional Experimental Results}

\vfive{
\subsection{Additional Baselines for General Domain Event Extraction}
\label{sec:general_domain_result_additional}

\mypar{Baseline selection criteria.} We select published EE models reporting performance on the ACE05 dataset using ED and EAE training data \textit{only} without using external resources (\eg knowledge graph) or additional tasks (\eg relation extraction, entity recognition) as our baselines for general domain EE experiments. We use the same data pre-processing, data splits and metrics as prior works \cite{wadden-etal-2019-entity,lin-etal-2020-joint}.

\mypar{Additional baselines.} In addition to the baselines we introduced in \secref{sec:experimental_setup}, we compare with \textbf{DyGIE++} \cite{wadden-etal-2019-entity}, a span graph-enhanced classification model for EE; \textbf{BERT\_QA} \cite{du-cardie-2020-event}, which formulates EE as an extractive question answering task with a sequence tagging classifier; \textbf{TANL} \cite{paolini2021structured}, which frames EE as a translation task between augmented natural languages; \textbf{BART-Gen} \cite{li-etal-2021-document}, which uses a sequence tagging model \cite{hou-etal-2020-shot} with additional keywords as input for ED and performs EAE by filling in event template with a conditional generation model; and \textbf{GTEE-DynPref} \cite{liu-etal-2022-dynamic}, which tunes dynamic prefix for generative EE models.

We do not compare with \citet{nguyen-etal-2022-learning,nguyen-etal-2021-cross,zhang-ji-2021-abstract} because they jointly learn additional tasks besides ED and EAE (\ie entity recognition and relation extraction) and there is no codebase provided by the time of this \vfinal{work}. We do not compare with \citet{wang-etal-2022-query} since its performance is worse than DEGREE \cite{hsu-etal-2022-degree} and GTEE-DynPref \cite{liu-etal-2022-dynamic} according to \citet{nguyen-etal-2022-learning}.

\begin{table}[ht!]
\begin{center}
\small
\setlength\tabcolsep{3pt}
\resizebox{\linewidth}{!}{
\begin{tabular}{cl|cc|ccHHHH}
\toprule
\multirow{2}{*}[-1pt]{\#} &
\multirow{2}{*}[-1pt]{Model} & \multicolumn{2}{c|}{10\%} & 
\multicolumn{2}{c}{100\%} 
\\
& & \multicolumn{1}{c}{Tri-C} & \multicolumn{1}{c|}{Arg-C} & \multicolumn{1}{c}{Tri-C} & \multicolumn{1}{c}{Arg-C}
\\
\midrule
1 & DyGIE++
& -- & 15.7\textsuperscript{\mathsection}
& 70.0\textsuperscript{\ddag} & 50.0\textsuperscript{\ddag} 
& -- & --
& -- & --
\\
2 & BERT\_QA 
& 50.1\textsuperscript{\ddag} & 27.6\textsuperscript{\ddag} 
& 72.4\textsuperscript{\dag} & 53.3\textsuperscript{\dag} 
& -- & --
& -- & --
\\
3 & TANL
& 54.8\textsuperscript{\ddag} & 29.0\textsuperscript{\ddag}
& 68.4\textsuperscript{\ddag} & 47.6\textsuperscript{\ddag} 
& 45.9 & 30.1
& 54.7 & 43.2
\\
4 & BART-Gen
& -- & --
& 71.1\textsuperscript{\dag} & 53.7\textsuperscript{\dag} 
& -- & --
& -- & --
\\
5 & GTEE-DynPref
& -- & --
& 72.6\textsuperscript{\dag} & 55.8\textsuperscript{\dag}
& -- & --
& -- & --
\\
\midrule
6 & Text2Event
& 47.0\textsuperscript{\ddag} & 24.9\textsuperscript{\ddag}
& 71.9\textsuperscript{\dag} & 53.8\textsuperscript{\dag}
& 42.4 & 28.7
& \textbf{59.4} & 48.3
\\
7 & OneIE
& 61.5\textsuperscript{\ddag} & 26.8\textsuperscript{\ddag}
& \underline{74.7}\textsuperscript{\dag} & \underline{56.8}\textsuperscript{\dag}
& 48.8 & 35.1
& 57.0 & 46.5
\\ 
8 & DEGREE
& \underline{66.1}\textsuperscript{\dag} & \underline{42.1}\textsuperscript{\dag}
& 72.2\textsuperscript{\dag} & 55.8\textsuperscript{\dag} 
& 50.6 & 39.4
& 57.8 & \textbf{50.4}
\\ \midrule
9 & \vfive{\modelname}
& \textbf{68.9} & \textbf{44.7}
& \textbf{75.5} & \textbf{57.6}
\\ 
\bottomrule
\end{tabular}
}
\caption{
\vtwo{
ED and EAE performance (F1 score, \%) on the general domain dataset ACE05. We report the numbers from the original paper (indicated by \dag), \cite{hsu-etal-2022-degree} (indicated by \ddag) and \cite{ye2022ontology} (indicated by \mathsection). \textbf{Boldface} denotes the best results while \ul{underscore} denotes the second best. \modelname achieves the best performance across both resource settings and tasks.
}
}
\label{table:overall_general_domain}
\end{center}
\end{table}
\mypar{Experimental results.} \tbref{table:overall_general_domain} shows the comparison with more baselines.
}

\subsection{Additional Ablation Studies}
\label{app:additional_ablation}
\vfive{
\mypar{Input prompt segments.}
We analyze the importance of prompt segments in \tbref{table:ablation_prompt}. For ED, we find that event type name is more important. For EAE, removing either the event type description (Line 5) or the argument role description (Line 9) leads to the most significant performance decreases. These results emphasize the benefits of incorporating the rich semantic information contained in the names and definitions for both event type and argument roles.
}

\begin{table}[th]
\begin{center}
\resizebox{\linewidth}{!}{
{
\small
\begin{tabular}{rl|ccc|ccc}
\toprule
\multirow{2}{*}{\#}
& \multirow{2}{*}{Prompt segments} 
& \multicolumn{3}{c|}{Identification} 
& \multicolumn{3}{c}{Classification} 
\\
& & P & R & F1 & P & R & F1
\\
\toprule
\multicolumn{8}{c}{Event Detection} 
\\
\midrule
1 & w/o type name &
71.19 & 63.32 & 67.02 & 67.41 & 60.73 & 63.90
\\
2 & w/o type description &
66.38 & 71.00 & 68.61 & 62.28 & 67.19 & 64.64
\\
3 & Vanilla \modelname
\vanillaDiceED
\\
\toprule
\multicolumn{8}{c}{Event Argument Extraction} 
\\
\midrule
4 & w/o type name
& 69.34 & 77.35 & 73.13 & 64.17 & 73.03 & 68.31
\\
5 & w/o type description
& 67.80 & 77.45 & 72.31 & 62.94 & 73.46 & 67.79
\\
6 & w/o trigger phrase
& 71.20 & 77.89 & 74.39 & 66.21 & 73.79 & 69.80
\\
7 & w/o trigger marker
& 68.55 & 78.53 & 73.20 & 64.70 & 75.51 & 69.69
\\
8 & w/o arg. role name
& 70.13 & 77.99 & 73.85 & 65.20 & 73.79 & 69.23
\\
9 & w/o role description
& 75.22 & 70.54 & 72.81 & 67.91 & 65.39 & 66.63
\\
10 & Vanilla \modelname
\vanillaDiceEAE
\\
\bottomrule
\end{tabular}
}
}
\caption{
Ablation study of prompt segments.
}
\label{table:ablation_prompt}
\end{center}
\end{table}

\mypar{Extraction vs typing formulation.} We formulate ED and EAE as conditional text generation tasks and consider two designs for our input and target format. The first is the \modelname design in which we expect the model to \textbf{extract} content given queries with event type/argument role information. 
%For ED, we query the model with a certain event type and expect the model to extract a sub-sequence of the input sentence as trigger text. Similarly, for EAE, we query the model with a certain argument role and expect model to extract argument text. 
The second design formulates a \textbf{typing} task that provides a query to the generative model for each mention so that the expected output is the predicted event type or argument role for the querying mention.
This approach is motivated by the notion that the output space of the typing formulation is much smaller than that of the extraction task. 
% For the ED example in \figref{fig:model}, we query with the input ``... calcified $<$query$>$plaque$<$/query$>$ ... artery. '' instead and the expected output is ``Event type is Sign\_symptom''.

We formulate the ED and EAE tasks as typing tasks by querying each possible mention. For the ED task, we first use the standalone mention identification module introduced in \secref{sec:components} to extract all possible triggers detected by the MI module, and then we query the generative model with the following example input and output format:

\begin{tcolorbox}
\small
Input: ... calcified \text{\color{blue}$<$query$>$}plaque\text{\color{blue}$<$/query$>$} ... artery. 
\\
Output: Event type is Sign\_symptom.
\end{tcolorbox}

The output is constrained to belong to the candidate pool of event types or the placeholder event type ``$<$Type$>$'' following the prefix ``Event type is ''. For the EAE task, we first extract all possible argument candidates and then query each candidate with the input sentence containing event trigger, event trigger marker, event type name and event type description:

\begin{tcolorbox}
\small
Input: ... densely \text{\color{blue}$<$query$>$}calcified\text{\color{blue}$<$/query$>$} \text{\color{blue}$<$trigger$>$}plaque\text{\color{blue}$<$/trigger$>$} ... artery. $\backslash$n Event type is Sign\_symptom. $\backslash$n Any symptom or clinical finding. $\backslash$n Event trigger is plaque.
\\
Output: Argument role is Detailed\_description.
\end{tcolorbox}

Similarly, the output is constrained to the candidate pool of argument roles possible for the given event type following the prefix ``Argument role is ''.

% We query with the following format for the example in \figref{fig:model}:

% \begin{tcolorbox}
% %%%\vspace{-0.5em}
% \small
% Input: ... calcified \text{\color{blue}$<$query$>$}plaque\text{\color{blue}$<$/query$>$} ... artery. 
% \\
% Output: Event type is Sign\_symptom.
% %%%\vspace{-0.5em}
% \end{tcolorbox}

% We compare 
%the vanilla \modelname extraction design with the typing design 
% the two formulations in \tbref{table:ablation_queryby}. 

\begin{table}[th]
\begin{center}
\resizebox{\linewidth}{!}{
{
\small
\setlength\tabcolsep{4pt}
\begin{tabular}{cl|ccc|ccc}
\toprule
\multirow{2}{*}{\#}
& \multirow{2}{*}{Formulation} 
& \multicolumn{3}{c|}{Identification} 
& \multicolumn{3}{c}{Classification} 
\\
& & P & R & F1 & P & R & F1
\\
\toprule
\multicolumn{8}{c}{Event Detection} 
\\
\midrule
1 & Typing
& 74.64 & 67.19 & 70.72 & 69.24 & 63.82 & 66.42
\\
2 & Extraction
\vanillaDiceED
\\
\toprule
\multicolumn{8}{c}{Event Argument Extraction} 
\\
\midrule
3 & Typing
& 58.59 & 44.95 & 50.87 & 53.63 & 41.14 & 46.56
\\
4 & Extraction
\vanillaDiceEAE
\\
\bottomrule
\end{tabular}
}
}
\caption{
Ablation study of generative task formulation. 
}
\label{table:ablation_queryby}
\end{center}
\end{table}
The results in \tbref{table:ablation_queryby} show that the typing formulation improves ED performance over extraction (though still worse than mention-enhanced \modelname), but leads to a much worse EAE performance. This is likely due to the typing task becoming more difficult as the number of candidate class increases 
%from ED to EAE task 
and complicated typing spaces varied by event types.

% \subsection{Experimental Result Variance}
% \label{app:variance}

% For results in \tbref{table:overall}, the variances for performances (\%) of \modelname for Tri-I, Tri-C, Arg-I and Arg-C are 0.19, 0.25, 0.21, 0.34. For results in \tbref{table:overall_general_domain_simple}, the variances for performances (\%) of \modelname for 10\% Tri-C, 10\% Arg-C, 100\% Tri-C, 100\% Arg-C are 0.30, 0.37, 0.24, 0.35.

\subsection{Full Low-Resource Results}
\label{app:full_low_resource_results}
We show the full low-resource experimental results illustrated in \figref{fig:downsample} in \tbref{table:downsample}.

\label{sec:appendix_downsample}
\begin{table}[h]
\begin{center}
\resizebox{\linewidth}{!}{
\setlength\tabcolsep{3pt}
\begin{tabular}{lH|
HHcccc|
HHcccc
}
\toprule
\multirow{2}{*}[-4pt]{Model} 
& 
& 1\% & 5\% & 10\% & 25\% & 50\% & 75\% & 1\% & 5\% & 10\% & 25\% & 50\% & 75\%
\\  \cmidrule{3-14}
& 
& \multicolumn{6}{c|}{Trigger Identification} 
& \multicolumn{6}{c}{Trigger Classification}
\\
\midrule
Text2Event & 
& & & -- & -- & -- & --
& & & 52.54 & 53.72 & 59.21 & 62.78
\\
OneIE & 
& & & 68.22 & 71.28 & 73.73 & 74.47
& & & 61.46 & 65.08 & 67.54 & 68.50
\\
DEGREE &
& & & 62.12 & 63.78 & 66.32 & 69.73
& & & 58.31 & 61.03 & 63.14 & 64.77
\\
\midrule
\modelname & 
& & & \textbf{71.47} & \textbf{72.79} & \textbf{74.07} & \textbf{74.88}
& & & \textbf{65.82} & \textbf{66.54} & \textbf{67.91} & \textbf{68.72}
\\ \toprule
& 
& \multicolumn{6}{c|}{Argument Identification} 
& \multicolumn{6}{c}{Argument Classification}
\\
\midrule
Text2Event & 
& & & -- & -- & -- & --
& & & 37.74 & 40.09 & 46.53 & 50.37
\\
OneIE & 
& & & 32.13 & 39.65 & 43.75 & 47.12 
& & & 24.95 & 32.36 & 35.70 & 38.55 
\\
DEGREE &
& & & 26.60 & 30.41 & 31.06 & 31.63
& & & 26.60 & 28.43 & 29.48 & 29.59
\\
\midrule
\modelname & 
& 00.00 & 00.00 & \textbf{49.97} & \textbf{53.55} & \textbf{54.42} & \textbf{55.83}
& 00.00 & 00.00 & \textbf{45.67} & \textbf{48.97} & \textbf{50.42} & \textbf{52.83}
\\
\bottomrule
\end{tabular}
}
\caption{
Performance on the downsampled training sets. We report the F1 score for each task using different downsampled training data. We create three random splits for each proportion and report the average performance.
}
\label{table:downsample}
\end{center}
\end{table}
\section{Details of Implementation and Experiments}
\label{sec:implementation_details}
\subsection{Implementation Details}
\mypar{Mention Identification.} The sliding window scans the passage from beginning to end with a pre-defined window size and step size, which significantly boosts the coverage of the predicted mentions. During both training and inference, we retain the original full-length input passage in addition to the sliding window segments.

\mypar{Training and evaluation.} We select the best epoch based on the highest F1 score of the most downstream MI/ED/EAE task on the validation set. When evaluating correctness, we only accept an exact match between the generated trigger/argument and the ground-truth trigger/argument as a correct prediction. We use beam search with 2 beams to generate the output sequences for all three generative tasks. The generation stops either when the ``end\_of\_sentence'' token is generated or the output length reaches 30. 

\mypar{Frameworks.} Our entire codebase is implemented in PyTorch.\footnote{\url{https://pytorch.org/}} The implementations of the transformer-based models are extended from the Huggingface\footnote{\url{https://github.com/huggingface/transformers}}~codebase~\cite{wolf-etal-2020-transformers}.

\subsection{Experiments Details}
% We report the averaged result for three runs with different random seeds for each experiment. 
We report the median result for five runs with different random seeds by default. 
For the low-resources result shown in \figref{fig:downsample}, we sample different selections of training data of corresponding proportion for each run.
% We report the averaged F1 score (y-axis) of three random splits for each proportion (x-axis). 
All the models in this work are trained on NVIDIA A6000 GPUs on a Ubuntu 20.04.2 operating system.

\subsection{Baseline Reproduction}
\label{appendix:baselines}
\mypar{Mention Identification.} For results in \tbref{table:overall_ET}, we use BART-large for \citeauthor{yan-etal-2021-unified-generative} because \citet{yan-etal-2021-unified-generative} only supports a generative model with absolute position embedding. OneIE uses BERT-large as its default and we use T5-large for our proposed \modelname-MI module.

\mypar{ED and EAE. } We use authors' codebases to produce baseline results. OneIE jointly learns ED, EAE, and MI tasks and we provide entity information to its MI module with event types and role types stripped to equate its training information with the training information provided to our model \modelname. For DEGREE, human-written templates that organize the argument roles of an event type in a sentence are required by the model. We construct these templates using phrases such as ``$<$Argument role$>$ is $<$argument text$>$'' for all potential argument roles of an event type as the template. 

\subsection{Hyperparameters}
For the ED module, we define positive instances as (\textsc{passage}, \textsc{event type}) pairs where the passage contains one or more event triggers of this event type. Negative instances are pairs in which the passage contains no event triggers of the event type. We create 10 negative instances for each positive instance.
For the EAE module, we define positive instance as the (\textsc{passage}, \textsc{event trigger}, \textsc{event type}, \textsc{argument role)} tuple that there exists an argument text contained in the passage that meets the query criteria. We create 10 negative instances for each positive instance.
For the MI module, we use a window size of 10 words, with a sliding step of 4 words. We retain the original full sequence in both training and evaluation.
We use an AdamW optimizer with a 1e-5 learning rate without gradient accumulation. 
% We search for the best hyperparameters according to the trigger/argument classification F1-score on the validation set and 
We show the hyperparameter search ranges and the final choices in \tbref{table:hyperparam}.
\begin{table*}[h]
\begin{center}
\resizebox{\linewidth}{!}{
{
\small
\setlength\tabcolsep{2pt}
\begin{tabular}{lll}
\toprule
Hyperparameter & Search Range & Best \\
\midrule
Negative instance \# for ED & 1, 2, 3, 4, 5, 8, 10, all & 10 \\
Negative instance \# for EAE & 1, 2, 3, 4, 5, 8, 10, all  & 10 \\
MI module sliding window size & 4, 6, 8, 10, 12 & 10 \\
MI module sliding window step & 2, 4, 6, 8, 10 & 4 \\
MI module sliding window retains original long sequence during training & True, False & True \\
MI module sliding window retains original long sequence during inference & True, False & False \\
Batch size & 1, 2, 3, 4 & 4\\
Learning rate & 1e-4, 5e-5, 1e-5, 5e-6, 1e-6 & 1e-5 \\
Decoding method & beam search, greedy & beam search \\
Max epochs & & 70 \\
\bottomrule
\end{tabular}
}
}
\caption{Hyperparameter search ranges and the best settings.}
\label{table:hyperparam}
\end{center}
\end{table*}

%Please add the following packages if necessary:
%\usepackage{booktabs, multirow} % for borders and merged ranges
%\usepackage{soul}% for underlines
%\usepackage[table]{xcolor} % for cell colors
%\usepackage{changepage,threeparttable} % for wide tables
%If the table is too wide, replace \begin{table}[!htp]...\end{table} with
%\begin{adjustwidth}{-2.5 cm}{-2.5 cm}\centering\begin{threeparttable}[!htb]...\end{threeparttable}\end{adjustwidth}

\begin{table*}[h]\centering
\begin{tabularx}{0.95\textwidth}{lX}
\toprule
Event Type & Role \\\midrule
Sign\_symptom & Biological\_structure, Detailed\_description, Severity, Lab\_value, Distance, Shape, Area, Color, Texture, Frequency, Volume, Quantitative\_concept, Qualitative\_concept, Biological\_attribute, Subject, Other\_entity, History, Mass\\
\hline
Diagnostic\_procedure & Lab\_value, Biological\_structure, Detailed\_description, Qualitative\_concept, Nonbiological\_location, Frequency,Distance, Subject, Shape, Quantitative\_concept, Texture, Severity, Age, Color, Area, Volume, Administration, Mass\\
\hline
Therapeutic\_procedure & Detailed\_description, Biological\_structure, Lab\_value, Dosage, Nonbiological\_location, Frequency, Distance,Qualitative\_concept, Subject, Quantitative\_concept, Area, Administration, Other\_entity\\
\hline
Disease\_disorder &Detailed\_description, Biological\_structure, Severity, Lab\_value, Quantitative\_concept, Distance, Nonbiological\_location, Shape, Volume, Qualitative\_concept, Area, Subject, Biological\_attribute\\
\hline
Medication & Dosage, Administration, Detailed\_description, Frequency, Lab\_value, Nonbiological\_location, Quantitative\_concept, Biological\_structure, Volume\\
\hline
Clinical\_event & Nonbiological\_location, Detailed\_description, Frequency, Biological\_structure, Subject, Lab\_value, Quantitative\_concept, Volume\\
\hline
Lab\_value & Biological\_structure, Detailed\_description, Color, Severity, Frequency\\
\hline
Activity & Detailed\_description, Nonbiological\_location, Biological\_structure, Other\_entity, Frequency, Lab\_value, Quantitative\_concept\\
\hline
Other\_event & Biological\_structure, Quantitative\_concept, Nonbiological\_location, Severity, Detailed\_description\\
\hline
Outcome & Nonbiological\_location, Subject, Detailed\_description, Age\\
\hline
Date & -\\
\hline
Time & -\\
\hline
Duration & -\\
\bottomrule
\end{tabularx}
\caption{Event types and corresponding argument roles in \dataname, the argument roles are ordered by their appearance frequency. The most appeared roles are listed first.}\label{table:event_ontology}
\end{table*}

\end{document}